\RequirePackage{tikz} \usetikzlibrary{calc, arrows.meta, intersections, patterns, positioning, shapes.misc, fadings, through,decorations.pathreplacing}

\documentclass[pdflatex,sn-mathphys]{sn-jnl}

\usepackage{adjustbox}
\usepackage{longtable}

\jyear{2022}%

\theoremstyle{thmstyleone}%
\theoremstyle{thmstyletwo}%

\theoremstyle{thmstylethree}%

\begin{document}

\title[Article Title]{Computational Sarcasm Analysis on Social Media: A Systematic Review}


\author[1]{\fnm{Faria} \spfx{Binte} \sur{Kader}}\email{faria@iut-dhaka.edu}
\equalcont{These authors contributed equally to this work.}

\author[1]{\fnm{Nafisa} \sur{Hossain Nujat}}\email{nafisa13@iut-dhaka.edu}
\equalcont{These authors contributed equally to this work.}

\author*[1]{\fnm{Tasmia} \spfx{Binte} \sur{Sogir}}\email{tasmia@iut-dhaka.edu}
\equalcont{These authors contributed equally to this work.}

\author[1]{\fnm{Mohsinul} \sur{Kabir}}\email{mohsinulkabir@iut-dhaka.edu}

\author[1]{\fnm{Hasan} \sur{Mahmud}}\email{hasan@iut-dhaka.edu}

\author[1]{\fnm{Kamrul} \sur{Hasan}}\email{hasank@iut-dhaka.edu}

\affil[1]{\orgdiv{Department of Computer Science and Engineering}, \orgname{Islamic University of Technology}, \orgaddress{\city{Dhaka}, \country{Bangladesh}}}


\abstract{Sarcasm can be defined as saying or writing the opposite of what one truly wants to express, usually to insult, irritate, or amuse someone. Because of the obscure nature of sarcasm in textual data, detecting it is difficult and of great interest to the sentiment analysis research community. Though the research in sarcasm detection spans more than a decade, some significant advancements have been made recently, including employing unsupervised pre-trained transformers in multimodal environments and integrating context to identify sarcasm. In this study, we aim to provide a brief overview of recent advancements and trends in computational sarcasm research for the English language. We describe relevant datasets, methodologies, trends, issues, challenges, and tasks relating to sarcasm that are beyond detection. Our study provides well-summarized tables of sarcasm datasets, sarcastic features and their extraction methods, and performance analysis of various approaches which can help researchers in related domains understand current state-of-the-art practices in sarcasm detection.}




\maketitle

\section{Introduction}\label{sec1}
The Cambridge Dictionary\footnote{\url{https://dictionary.cambridge.org/}} defines sarcasm as the use of remarks that clearly mean the opposite of what they say, made in order to hurt someone's feelings or to criticize something in a humorous way. Sarcastic remarks, being very different in literal meaning than the actual intention of the speaker, is often difficult to interpret. Sarcasm detection is a form of sentiment analysis where a given statement is classified as either sarcastic or non-sarcastic. Textual sarcasm differs from sarcasm in verbal or face-to-face conversations. In textual data, sarcasm detection is significantly more challenging due to the absence of cues like body language, facial expressions, vocal tone, etc. Sarcasm detection is an important task because distinguishing sarcastic texts from normal ones improves the sentiment analysis performance (Gupta et al. \cite{gupta2017crystalnest}) by capturing the real sentiment of the given sentence. \\

Understanding sarcasm has long piqued the interest of researchers in the disciplines of psychology and sociology \cite{gibbs1986psycholinguistics,slugoski1988cruel,ball1965sarcasm}. At the beginning of the 21st century, computational sarcasm recognition piqued the interest of the natural language processing community. From the oldest known work by Tepperman et al. \cite{tepperman2006yeah}, the task of automatic sarcasm detection has come a long way. Starting with recognizing sarcasm from simple textual and vocal features, over the years, studies have been conducted using visual, numerical and even multi-modal data. While Pattern, Rule-based, and Statistical techniques dominated the research field in the first decade, various deep learning and transformer-based models have gained prominence in recent years. Given its popularity, this comes as no surprise that several works have been done analysing the studies under this domain. A recent survey by Yaghoobian et al. \cite{yaghoobian2021sarcasm} focused on the computational attempts at automatic sarcasm detection: the use of hashtag to annotate datasets, semi-supervised pattern extraction to identify implicit sentiment and the utilization of context. Another study by Sarsam et al. \cite{sarsam2020sarcasm} reviewed sarcasm detection using machine learning algorithms in Twitter, the major sarcasm classifiers, their classification performance and the features contributing to such performance. Chaudhari et al. \cite{chaudhari2017literature} did a literature survey on sarcasm detection where they briefly discuss about types of sarcasm, approaches for sarcasm detection and some issues and challenges. Wicana et al. \cite{wicana2017review} also presented a review on sarcasm detection from machine-learning perspective. These studies mostly focused on a single type of sarcasm detection algorithm or reviewed various literature from a specific aspect. Joshi et al. \cite{joshi2016automatic} provided a thorough review on automatic sarcasm detection: datasets, approaches, reported performance, trends and issues etc. However, there has been a recent surge in sarcasm detection research due to the introduction of new trends and techniques in this sector, such as the usage of transformer-based models. Through our study, we aim to construct a complete guide to the different aspects of sarcasm detection in social media. We focused on the trends and how changes in methodologies and features used in sarcasm detection throughout the years have affected the classification performance. Additionally, we presented a comparative study on different datasets and features and discussed the issues and challenges observed throughout recent sarcasm detection research. We believe our work will be a helpful guide in seeing the complete picture of sarcasm detection in social media. Throughout this study, we also tried to answer the following research questions -
\begin{enumerate}
\item What type of features and datasets are relevant and mostly used in detecting sarcasm on social media?
\item What advancements have been made in sarcasm detection methodologies throughout the years? 
\item What are the issues and challenges that are currently in the way of improving sarcasm detection performance in social media? 
\item Except for detection, what research has been done on sarcasm?\\
\end{enumerate} 

The remaining sections of the paper are structured as follows: section \ref{sec2} describes the notion sarcasm detection, section \ref{sec4} reviews different datasets and their types while section \ref{sec5} reviews types of features and their extraction methods, section \ref{sec6} explores methodologies as well as shared tasks for sarcasm detection, section \ref{sec7} presents the recent trends in this domain, and issues and challenges of sarcasm detection are discussed in section \ref{sec8}.

\section{Sarcasm Detection}\label{sec2}
This section briefly defines sarcasm, its types, sarcasm detection and the comparison between irony and sarcasm.

\subsection{Definition}\label{sec2.1}
Sarcasm is a negative sentiment hidden beneath a positive surface. It is different from lying as deception is not a part of the intention of its speaker. Its main goal is to express humorous analysis or commentary (Nakassis et al. \cite{nakassis2002beyond}). According to Tepperman et al. \cite{tepperman2006yeah}, sarcasm is the name given to speech bearing a semantic interpretation exactly opposite to its literal meaning. Poria et al. \cite{poria2016deeper} defines sarcasm as a sharp, bitter, or cutting expression or remark; a bitter gibe or taunt. However, sarcasm is a deeper concept, highly related to the language, and to the common knowledge (Bouazizi et al. \cite{bouazizi2016pattern}). Ivanko et al. \cite{ivanko2003context} proposed a 6-tuple sarcasm representation consisting of a speaker, a listener, a context, an utterance, a literal proposition and intended proposition. \\

Sarcasm detection is a classification task in its most typical form. From a given text, the task includes classifying the text as sarcastic or non-sarcastic. Sarcasm detection is a fairly recent but promising research field in the domain of Natural Language Processing. Nonetheless, it serves as a crucial part to sentiment analysis (Maynard et al. \cite{maynard2014cares}). In the following section, sarcasm and its various forms are discussed in detail.

\subsection{Types of Sarcasm}\label{sec2.2}
Various works tried to classify sarcasm further into different types. Abulaish et al. \cite{abulaish2018self} mentions seven categories of sarcasm: self-deprecating, brooding, deadpan, polite, obnoxious, manic, and raging. Sundararajan et al. \cite{sundararajan2020multi} classified sarcasm into four types namely Polite, Rude, Deadpan, and Raging. A few works only focused on self-deprecating sarcasm detection (Kamal et al. \cite{kamal2019lstm}) which are self-referential sarcasm. Oprea et al. \cite{oprea2019exploring} suggested that there are two types of sarcasm: intended and perceived, and both of the types should be treated as separate phenomena. They worked with two types of datasets where one was manually labeled (perceived) and another one was annotated via distant supervision (intended). The performance on the manually labeled dataset was poor as the annotators may not have understood the true intentions of the authors.

\subsection{Irony vs Sarcasm}\label{sec2.3}
We observed many works stating that they did not differentiate between sarcasm and irony because they are not so easily distinguishable even for human experts \cite{ilic2018deep,dimovska2018sarcasm,potamias2020transformer,naseem2020towards}. However, a few works across various social media platforms attempted to distinguish sarcasm from irony. In literature, we can notice a subtle difference between sarcasm and irony. When something appears to be going against your expectations, it is said to be ironic. If an expectation is black, then an ironic outcome would be white, not off-white or gray\footnote{\url{https://www.vocabulary.com/articles/chooseyourwords/irony-satire-sarcasm/}}. On the contrary, sarcasm is typically used negatively because it is frequently directed at a specific individual and appears as a witty mockery\footnote{\url{https://grammar.yourdictionary.com/vs/irony-vs-sarcasm-types-and-differences.html}}. Ling et al. \cite{ling2016empirical} tried to find out the latent structural difference between ironic and sarcastic tweets. Khokhlova et al. \cite{khokhlova2016distinguishing} differentiated between sarcasm and ironic tweets based on use of hashtags, structure of tweets, ratios of parts of speech, frequency of words and phrases, lastly compared with the NRC Word-Emotion Association Lexicon (EmoLEx)\footnote{\url{https://colab.research.google.com/github/littlecolumns/ds4j-notebooks/blob/master/upshot-trump-emolex/notebooks/NRC\%20Emotional\%20Lexicon.ipynb}} and proposed that sarcastic tweets may seem more positive than ironic tweets. A shared task was proposed on irony detection in SemEval-2018\footnote{\url{https://github.com/Cyvhee/SemEval2018-Task3/tree/master/datasets}} which was the 12\textsuperscript{th} workshop on semantic evaluation where the dataset contained tweets collected using irony-related hashtags (i.e. \#irony, \#sarcasm, \#not). Dimovska et al. \cite{dimovska2018sarcasm} demonstrated the impact of various features on irony and sarcasm detection separately in the SemEval-2018 dataset and their best performing irony detection model was a linear SVM using the hashing vectorizer on the word n-grams. 

\section{Search Strategy and Inclusion Criteria}\label{sec3} 
In this section, we briefly describe the process of collecting research papers which worked with sarcasm detection. In their survey, Joshi et al. \cite{joshi2016automatic} thoroughly reviewed different works related to sarcasm detection up until 2015 but there has been a surge of new work in the field since then. Although there has been many review works as well, they were mostly focused on specific technologies and trends. With an aim to provide a thorough discussion on sarcasm detection as a whole, we focused on the studies since the year 2016 and onwards. We looked for relevant papers with keywords like ‘Sarcasm Detection’ and also went through references of notable sarcasm detection studies. We excluded papers based on the following criteria to have a reduced scope of survey-

\begin{enumerate}
\item We included studies that worked with social media datasets. This criterion excluded papers which only worked with datasets relating to online news portals, scripts from TV series, Movies, Books etc.
\item We mainly focused on the works which used English datasets. Studies not containing any English dataset were thus excluded.  
\item Studies with only textual-data or multi-modal data with textual data being one of the modes were prioritized. We excluded any research that worked with only images, video or audio data.
\end{enumerate}

\section{Datasets}\label{sec4}
This section describes datasets used in different sarcasm detection studies. Most of these studies train and test on already available popular datasets such as the datasets used by Riloff et al. \cite{riloff2013sarcasm}, Khodak et al. \cite{khodak2017large} and Cai et al. \cite{cai2019multi}. We observed that Twitter is predominantly the most popular social media platform used for sarcasm detection datasets although Reddit, Amazon and a few discussion forums were also seen being used. These datasets can be divided into three main categories namely short text, long text and image data types.

\subsection{Short Text}\label{sec4.1}
Short texts are by far the most popular form of datasets observed throughout the research done in sarcasm detection. The social media datasets from studies we focused on are primarily short texts due to most social media platforms having a length limit for posts and comments. These short texts are mostly found to be acquired from twitter and reddit. Twitter, the social networking site, is known as a microblogging platform for its character limit of 280. With its 330 million monthly active users\footnote{\url{https://financesonline.com/number-of-twitter-users/}} ranging from teenagers to elderly people, Twitter is an excellent platform for gathering data relating to sarcasm and irony. We found that the Twitter API\footnote{\url{https://developer.twitter.com/en/docs/twitter-api}} is widely used for data acquisition. Riloff et al. \cite{riloff2013sarcasm} implemented a twitter dataset that contains 3,200 tweets, out of which 742 are labeled as sarcastic and 2,458 as non-sarcastic. The dataset is later seen in many significant sarcasm detection works \cite{joshi2016automatic,ghosh2016fracking,tay2018reasoning}. Reddit is another such platform allowing slightly larger sized contents. Reddit can also be categorized under the short text category as it still has a length limit, unlike most discussion forums with no limits defined (Joshi et al. \cite{joshi2016automatic}). Reddit has over 430 million monthly active users\footnote{\url{https://earthweb.com/how-many-people-use-reddit/}} who are mostly younger people and thus provide a great source of data for sarcasm detection. One of the most popular sarcasm detection dataset is proposed by Khodak et al. \cite{khodak2017large} known as SARC (Self-Annotated Reddit Corpus)\footnote{\url{https://nlp.cs.princeton.edu/SARC/}}. The SARC dataset contains 1.3 million sarcastic and 532 million non-sarcastic posts from Reddit. This dataset is later found in other research; one such work is by Hazarika et al. \cite{hazarika2018cascade}. Apart from these, there are a lot of other datasets containing mainly short texts, either obtained as a subset of these two datasets or created with new Twitter or Reddit data. The Twitter and Reddit dataset was utilized in experiments for the SemEval-2018 (Semantic Evaluation 2018) shared task, such as in the works of Ilic et al. \cite{ilic2018deep} and Wu et al. \cite{wu2018thu_ngn}. We observed a few other forms of short text data collection from book snippets, online comments etc. \cite{joshi2016word,bharti2016sarcastic}. 

\subsection{Long Text}\label{sec4.2}
Long texts are the second most popular form of datasets observed in sarcasm detection research. Most of these datasets contain product reviews from Amazon \cite{dharwal2017automatic,agrawal2018affective,parde2018detecting}. Amazon, being the biggest e-commerce platform, has millions of products and hundreds of reviews for these products. Filatova et al. \cite{filatova2012irony} created a dataset containing 437 sarcastic reviews and 817 regular reviews from Amazon. Another such dataset was developed by Mishra et al. \cite{mishra2019modular}. Although other types and forms of data were included in it as well. Discussion forums are also a popular source of long text data. Oraby et al. \cite{oraby2017you} created a dataset with 2496 sarcastic and non-sarcastic remarks from debate forums. We observed that these discussion forum data are typically used alongside other forms of data from different social media platforms. One such dataset was seen in the work of Bharti et al. \cite{bharti2016sarcastic} with data from Twitter, product reviews, comments, books and discussion forums. News portals, Facebook posts and yelp reviews are also good sources of long text sarcasm data. Subramanian et al. \cite{subramanian2019exploiting} used both Twitter and facebook datasests. Use of datasets with long texts is a fairly recent trend as a result of different e-commerce or review sites and web portals other than twitter and reddit gaining popularity in the last few years.

\subsection{Image}\label{sec4.3}
As we have mostly focused on textual datasets with some cases of multimodal datasets with text data as one of the modals, we found some image data where texts are attached with specific images as captions. Most of the multimodal datasets that we focused on are created with tweets containing both images and texts. For example Cai et al. \cite{cai2019multi} implemented a multimodal dataset\footnote{\url{https://github.com/headacheboy/data-of-multimodal-sarcasm-detection}} with 14075 sarcastic and 10560 non-sarcastic tweets including images. This dataset was later seen in many other research \cite{wang2020building,pan2020modeling,xu2020reasoning}. Another popular multimodal dataset was created by Schifanella et al. \cite{schifanella2016detecting} with texts and images from Instagram, Tumblr and Twitter. \\

Table \ref{tab1} shows the summary of some of the datasets used in different sarcasm detection works. The columns under annotation section defines the type of annotation used in the corpus: manual annotation, annotation using hashtags or unlabeled dataset. Many research might use a combination of datasets which have several type of annotations \cite{ptavcek2014sarcasm,poria2016deeper,oraby2017you}.

\begin{center}
\begin{longtable}{|p{0.16\textwidth}|p{0.03\textwidth}|p{0.02\textwidth}|p{0.03\textwidth}|p{0.16\textwidth}|p{0.16\textwidth}|p{0.03\textwidth}|p{0.03\textwidth}|p{0.03\textwidth}|}
\caption{Summary of sarcasm detection datasets from different social media platforms}
\label{tab1} \\
\hline
\textbf{}&\multicolumn{3}{|c|}{\textbf{Dataset}}&\textbf{}&\textbf{}&\multicolumn{3}{|c|}{\textbf{Annotation}}\\
\hline
\textbf{} 
& \begin{sideways}\textbf{Short Text}\end{sideways}
& \begin{sideways}\textbf{Long Text}\end{sideways}
& \begin{sideways}\textbf{Image}\end{sideways}
& \textbf{Samples}
& \textbf{Platform}
& \begin{sideways}\textbf{Manual}\end{sideways}
& \begin{sideways}\textbf{Hashtag}\end{sideways}
& \begin{sideways}\textbf{None}\end{sideways}
\\
\hline

Filatova et al. \cite{filatova2012irony} & & \checkmark & & 1254 & Amazon & \checkmark & &\\
\hline
Riloff et al. \cite{riloff2013sarcasm} & \checkmark & & & 1600 & Twitter & \checkmark & &\\
\hline
Ptáček et al. \cite{ptavcek2014sarcasm} & \checkmark & & & 920000 & Twitter & \checkmark & \checkmark &\\
\hline
Barbieri et al. \cite{barbieri2014modelling} & \checkmark & & & 60000 & Twitter & & \checkmark &\\
\hline
Bamman et al. \cite{bamman2015contextualized} & \checkmark & & & 19534 & Twitter & & \checkmark &\\
\hline
Amir et al. \cite{amir2016modelling} & \checkmark & & & 11541 & Twitter & & \checkmark &\\
\hline
Bharti et al. \cite{bharti2016sarcastic} & \checkmark & & & 1.5M & Twitter & & & \checkmark\\
\hline
Joshi et al. \cite{joshi2016word} & \checkmark & & & 3629 & Goodreads & & \checkmark &\\
\hline
Ghosh et al. \cite{ghosh2016fracking} & \checkmark & & & 41000 & Twitter & & \checkmark &\\
\hline
Poria et al. \cite{poria2016deeper} & \checkmark & & & 100000 & Twitter & \checkmark & \checkmark &\\
\hline
Schifanella et al. \cite{schifanella2016detecting} & \checkmark & & \checkmark & 600925 & Instagram, Tumblr, Twitter & & \checkmark &\\
\hline
Zhang et al. \cite{zhang2016tweet} & \checkmark & & & 9104 & Twitter & & \checkmark &\\
\hline
Felbo et al. \cite{felbo2017using} & \checkmark & & & 1.6B & Twitter & & & \checkmark\\
\hline
Ghosh et al. \cite{ghosh2017magnets} & \checkmark & & & 41200 & Twitter & \checkmark & &\\
\hline
Khodak et al. \cite{khodak2017large} & \checkmark & & & 533.3M & Reddit & \checkmark & &\\
\hline
Oraby et al. \cite{oraby2017you} & & \checkmark & & 10270 & Debate forum & \checkmark & \checkmark &\\
\hline
Prasad et al. \cite{prasad2017sentiment} & \checkmark & & & 2000 & Twitter & \checkmark & &\\
\hline
Baziotis et al. \cite{baziotis2018ntua} & \checkmark & & & 550M & Twitter & & & \checkmark\\
\hline
Hazarika et al. \cite{hazarika2018cascade} & \checkmark & & & 219368 & Reddit & \checkmark & &\\
\hline
Ghosh et al. \cite{ghosh2018sarcasm} & \checkmark & \checkmark & & 36391 & Twitter, Reddit, Discussion Forum & \checkmark & \checkmark &\\
\hline
Ilic et al. \cite{ilic2018deep} & \checkmark & \checkmark & & 419822 & Twitter, Reddit, Debate Forum & \checkmark & \checkmark &\\
\hline
Tay et al. \cite{tay2018reasoning} & \checkmark & \checkmark & & 94238 & Twitter, Reddit, Debate Forum & \checkmark & \checkmark &\\
\hline
Van et al. \cite{van2018semeval} & \checkmark & & & 4792 & Twitter & \checkmark & \checkmark &\\
\hline
Wu et al. \cite{wu2018thu_ngn} & \checkmark & & & 4618 & Twitter & \checkmark & \checkmark &\\
\hline
Majumder et al. \cite{majumder2019sentiment} & \checkmark & & & 994 & Twitter & & \checkmark &\\
\hline
Cai et al. \cite{cai2019multi} & & & \checkmark & 24635 & Twitter & & \checkmark &\\
\hline
Kumar et al. \cite{kumar2019sarcasm} & \checkmark & \checkmark & & 24635 & Twitter, Reddit, Debate Forum & & \checkmark &\\
\hline
Subramanian et al. \cite{subramanian2019exploiting} & \checkmark & \checkmark & & 12900 & Twitter, Facebook & & \checkmark &\\
\hline
Jena et al. \cite{jena2020c} & \checkmark & & & 13000 & Twitter, Reddit & \checkmark & \checkmark &\\
\hline
Potamias et al. \cite{potamias2020transformer} & \checkmark & & & 533.3M & Twitter, Reddit & \checkmark & \checkmark &\\
\hline

\end{longtable}
\end{center}

\section{Features}\label{sec5}
This section describes the types of features used in the experiments in concern and their various extraction methods.

\subsection{Types of Features}\label{sec5.1}
Prior to the rise of deep learning-based models, handpicked features were generated from texts in order to detect sarcasm in social media texts. The features can be classified into a few major groups, which are briefly outlined in this section.

\subsubsection{Lexical}\label{sec5.1.1}
We observed that lexical features are by far the most used features in sarcasm detection in social media. Types of features which fall under the lexical kind are characters, n-grams, sentences, numbers, hashtags etc. N-grams are widely used features in all types of natural language processing research, most of them being unigrams and bigrams \cite{ling2016empirical,joshi2016word,schifanella2016detecting}. Many studies also use character n-grams as well as full sentences as features, one such example can be seen in the work of Dimovska et al. \cite{dimovska2018sarcasm} who used four different groups of features: character unigrams, character n-grams (where n was set between 1 and 4), word unigrams and word n-grams (where n was set between 1 and 3). Some experiments also include numerical features \cite{kumar2017having,dubey2019numbers}. Hashtags have also been used as features as they can provide insight into sarcastic intentions \cite{ghosh2016fracking,ilic2018deep}.

\subsubsection{Pragmatic}\label{sec5.1.2}
Pragmatic features are usually expressions and reactions used in text data. For example emoticons and smileys are very popular ways of expressing feelings which can be used to differentiate between sarcastic and non-sarcastic utterances. Bharti et al. \cite{bharti2016sarcastic} used emoticons and smileys as a part of their feature sets. Similarly, ratings and reactions associated with social media contents can also indicate sarcasm and irony. Pragmatic features are thus seen in many social media sarcasm detection works. Das et al. \cite{das2018sarcasm} used six types of user reaction count related to a post on Facebook as one of their features. Parde et al. \cite{parde2018detecting} incorporated Amazon star ratings in their feature set. Felbo et al. \cite{felbo2017using} included pragmatic features alongside lexical features in their experiment. In addition to word-embedding based feature sets, Onan et al. \cite{onan2019topic} considered pragmatic features along with lexical, implicit incongruity and explicit incongruity based features.

\subsubsection{Hyperbole}\label{sec5.1.3}
Hyperbole features such as interjections, intensifiers, punctuation are also important in sarcasm detection as they can help understand the relationships among words and other features as well as the level of importance of a sentence. Kumar et al. \cite{kumar2019sarcasm} considered the following five punctuation-based features: number of exclamation marks, number of question marks, number of periods, number of capital letters and number of uses of ``or" in the tweets. Another type of hyperbole feature is capitalization which can mean emphasis on certain n-grams and thus can be an important feature. One example of capitalization as a feature can be seen in the feature set of Prasad et al. \cite{prasad2017sentiment}.

\subsubsection{Semantic}\label{sec5.1.4}
Average length of words, number of times a certain word is used, length of a sequence etc. are types of semantic features. These are used as additional information about a statement. Chakrabarty et al. \cite{chakrabarty2020r} used semantic incongruity as a feature in their experiment.

\subsubsection{Syntactic}\label{sec5.1.5}
Amir et al. \cite{amir2016modelling} mentioned syntactic features in their experiment among several other features. Use of Parts-of-Speech (POS) tags are very popular resulting from their ability to signify the type of words or tuples. Ghosh et al. \cite{ghosh2016fracking} used Parts-of-Speech tags for sarcasm detection. The extraction method used was POS tagging which is the process of converting a sentence into words or tuples and then associating tags to them\footnote{\url{https://www.geeksforgeeks.org/nlp-part-of-speech-tagged-word-corpus/}}.

\subsubsection{Sentiment}\label{sec5.1.6}
Sentiment features can be the polarity or the emotional level of a statement and are considered useful in identifying sarcasm. Khokhlova et al. \cite{khokhlova2016distinguishing} mentioned sentiment polarity as one of the features in their experiment. Sarcasm and irony are ways to evoke a certain sentiment in a person and that is why sentiment is seen being used as an important type of feature in sarcasm detection. Several sarcasm detection studies were found which mentioned sentiment as a type of feature \cite{joshi2016word,ghosh2016fracking,amir2016modelling}. Poria et al. \cite{poria2016deeper} concluded that sentiment and emotion features are the most useful features, besides baseline features in their experiment. Schifanella et al. \cite{schifanella2016detecting} extracted subjectivity and sentiment scores as features from their multimodal dataset and fed into their sarcasm detection model.

\subsubsection{Context}\label{sec5.1.7}
Contextual features have been very popular in recent years. Use of contextual features drastically eased the process of sarcasm detection and many studies have used them since \cite{amir2016modelling,ghosh2017magnets,ghosh2017role,sreelakshmi2018effective,poria2016deeper}. Both the Twitter and Reddit datasets provided in the FigLang2020\footnote{\url{https://competitions.codalab.org/competitions/22247}} shared task have conversational contexts between the users and the respective responses, that are supposed to be classified as Sarcastic or Non Sarcastic responses, using the contexts. These contextual features can be author or addressee information, audience, response, environment, history etc. Hazarika et al. \cite{hazarika2018cascade} utilized both content and contextual-information by processing contextual information with the use of user profiling to create user embeddings that capture indicative behavioral traits for sarcasm. Zhang et al. \cite{zhang2016tweet} used local and contextual features in their sarcasm detection study and showed that using only local tweet features, the neural model achieves an accuracy of 78.55\% whereas using local and  contextual features together increases the accuracy of the neural model upto 90.74\%, which proves the significance of contextual features in sarcasm detection.

\subsubsection{Image}\label{sec5.1.8}
As mentioned in the Datasets section (section \ref{sec4}), we’ve mostly focused on textual datasets in this study with a few cases of multimodal datasets where texts are attached with specific images as captions. Studies with multimodal datasets such as the popular dataset by Cai et al. \cite{cai2019multi} mainly used two feature groups namely text features and image features. Image features can be a great insight into the context and meaning of associated texts. Schifanella et al. \cite{schifanella2016detecting} adapted a visual neural network initialized with parameters trained on ImageNet\footnote{\url{https://www.image-net.org/}} to multimodal (text+image) sarcastic posts and concluded that visual features boost the performance of the textual models.

\subsection{Feature Extraction}\label{sec5.2}
The most popular feature extraction methods observed in this study are Bag-of-Words (BoW) and Term Frequency-Inverse Document Frequency (TF-IDF). Bag-of-Words method essentially transforms a document into a collection of words and is the simplest form of feature extraction in textual data. TF-IDF introduces two new terms to the BoW method by assigning weights to words inside the documents\footnote{\url{https://www.geeksforgeeks.org/feature-extraction-techniques-nlp/}}. Ghosh et al. \cite{ghosh2017role}, Xiong et al. \cite{xiong2019sarcasm}, Jamil et al. \cite{jamil2021detecting} incorporated the Bag-of-Words feature extraction technique in their studies. In their experiment, Zhang et al. \cite{zhang2016tweet} used TF-IDF to extract some of the features in their feature sets. Dharwal et al. \cite{dharwal2017automatic}, Jain et al. \cite{jain2017sarcasm}, Onan et al. \cite{onan2019topic} also used TF-IDF as one of the feature extraction techniques in their respective research. However, these two techniques have their limitations. Focusing on the frequency of occurrence, both BoW and TF-IDF fail to grasp the context of a text which might be very crucial for sarcasm detection. \\

Different Word Embedding techniques have also been used as ways to transform words into vectors. Word2Vec is a very popular word embedding technique. Word2Vec uses unsupervised learning where words are mapped with the ones they occur the most with. To extract features from their multimodal dataset, Schifanella et al. \cite{schifanella2016detecting} used Word2Vec technique. Joshi et al. \cite{joshi2016word}, Oraby et al. \cite{oraby2017you} also used Word2Vec technique for feature extraction. There are mainly two Word2Vec techniques namely Continuous Bag of Words (CBoW) and Skip-Gram\footnote{\url{https://www.geeksforgeeks.org/word-embeddings-in-nlp/}}. The CBOW method predicts a target word based on context whereas the Skip-Gram method predicts a target word using the words adjacent to it\footnote{\url{https://fasttext.cc/docs/en/unsupervised-tutorial.html}}. There also are a few variations of Word2Vec such as Doc2Vec, Emoji2Vec (Eisner et al. \cite{eisner2016emoji2vec}) etc. which use a similar idea to generate vectors from different types of corpus. Khotijah et al. \cite{khotijah2020using} used Doc2Vec technique to extract features while Subramanian et al. \cite{subramanian2019exploiting} extracted and embedded emoji tokens using Emoji2Vec technique. Another word embedding technique is GloVe (Global Vectors)\footnote{\url{https://nlp.stanford.edu/projects/glove/}} where the model efficiently leverages statistical information by training only on the nonzero elements in a word-word co-occurrence matrix, rather than on the entire sparse matrix or on individual context windows in a large corpus and produces a vector space with meaningful substructure (Pennington et al. \cite{pennington2014glove}). Cai et al. \cite{cai2019multi} performed modality fusion in their multimodal dataset instead of simply concatenating the feature vectors from different modalities and made use of GloVe technique to attain those vectors. FastText is also one of the most frequently used Word Embedding techniques. The FastText algorithm is pretty similar to Word2Vec but it uses N-grams along with the collection of words and thus generates different variations of words\footnote{\url{https://towardsdatascience.com/word-embedding-techniques-word2vec-and-tf-idf-explained-c5d02e34d08}}. Mehndiratta et al. \cite{mehndiratta2019identification} used Word2Vec, GloVe and FastText techniques to extract features. Onan et al. \cite{onan2019topic} also used those three feature extraction techniques. Although Word2Vec and Glove word embedding techniques work very well in mapping and labeling data, they place words close to each other which are actually opposite in meaning, thus making sarcasm detection difficult. They also struggle with out of vocabulary words, though FastText partially addresses them by using n-grams.. \\

There are many popular machine learning models that are also being used to aid the feature extraction process such as Convolution Neural Network (CNN), Support Vector Machine (SVM), many variations of Bidirectional Encoder Representations from Transformers (BERT) and Long Short-Term Memory (LSTM), Embeddings from Language Models (ELMo) etc. Bharti et al. \cite{bharti2016sarcastic} performed a Hidden Markov model or HMM-based algorithm for POS tagging. Many studies used the NRC Word-Emotion Association Lexicon (EmoLEx)\footnote{\url{https://colab.research.google.com/github/littlecolumns/ds4j-notebooks/blob/master/upshot-trump-emolex/notebooks/NRC\%20Emotional\%20Lexicon.ipynb}}. This lexicon is a list of 14,182 English words (unigrams) that belong to two sentiments (positive or negative) and are labeled with eight Plutchik’s primary emotions (anger, anticipation, disgust, fear, joy, sadness, surprise, trust) (Khokhlova et al. \cite{khokhlova2016distinguishing}). Agrawal et al. \cite{agrawal2018affective} computed sentiment labels using EmoLex in their experiment. Some of the other significant feature extraction methods are ResNet, SentiWordNet (Baccianella et al. \cite{baccianella2010sentiwordnet}), SentiBank (Borth et al. \cite{borth2013sentibank}), TExtBlob\footnote{\url{https://textblob.readthedocs.io/en/dev/}}, LIWC\footnote{\url{https://www.liwc.app/}}, COMET (Bosselut et al. \cite{bosselut2019comet}) etc. \\

Table \ref{tab2} shows the summary of a few types of features and feature extraction methods used in different sarcasm detection works.

\begin{center}
\begin{longtable}{|p{0.16\textwidth}|p{0.02\textwidth}|p{0.02\textwidth}|p{0.02\textwidth}|p{0.02\textwidth}|p{0.02\textwidth}|p{0.02\textwidth}|p{0.02\textwidth}|p{0.02\textwidth}|p{0.02\textwidth}|p{0.02\textwidth}|p{0.02\textwidth}|p{0.02\textwidth}|p{0.02\textwidth}|p{0.02\textwidth}|}
\caption{Summary of types of features and feature extraction methods in sarcasm detection}
\label{tab2} \\
\hline
\textbf{}&\multicolumn{8}{|c|}{\textbf{Type of Feature}}&\multicolumn{6}{|c|}{\textbf{Extraction Method}} \\
\hline
\textbf{} 
& \begin{sideways}\textbf{Lexical}\end{sideways}
& \begin{sideways}\textbf{Pragmatic}\end{sideways}
& \begin{sideways}\textbf{Hyperbole}\end{sideways}
& \begin{sideways}\textbf{Semantic}\end{sideways}
& \begin{sideways}\textbf{Syntactic}\end{sideways}
& \begin{sideways}\textbf{Sentiment}\end{sideways}
& \begin{sideways}\textbf{Context}\end{sideways}
& \begin{sideways}\textbf{Image}\end{sideways}
& \begin{sideways}\textbf{BoW}\end{sideways}
& \begin{sideways}\textbf{TF-IDF}\end{sideways}
& \begin{sideways}\textbf{Word2vec}\end{sideways}
& \begin{sideways}\textbf{GloVe}\end{sideways}
& \begin{sideways}\textbf{FastText}\end{sideways}
& \begin{sideways}\textbf{Machine Learning Models}\end{sideways}
\\
\hline

Amir et al. \cite{amir2016modelling} & \checkmark & & \checkmark & & \checkmark & \checkmark & \checkmark & & \checkmark & & & & & \checkmark
\\
\hline
Bharti et al. \cite{bharti2016sarcastic} & & & & & \checkmark & \checkmark & & & & & & & & \checkmark
\\
\hline
Ghosh et al. \cite{ghosh2016fracking} & & & & & \checkmark & \checkmark & & & \checkmark & & & & & \checkmark \\
\hline
Joshi et al. \cite{joshi2016word} & \checkmark & \checkmark & \checkmark & \checkmark & & \checkmark & & & & & \checkmark & \checkmark & & \\
\hline
Poria et al. \cite{poria2016deeper} & & & & \checkmark & & \checkmark & & & & & & & & \checkmark \\
\hline
Schifanella et al. \cite{schifanella2016detecting} & \checkmark & & & \checkmark & & \checkmark & & \checkmark & & & \checkmark & & & \checkmark \\
\hline
Zhang et al. \cite{zhang2016tweet} & \checkmark & & & & & & \checkmark & & & \checkmark & & & & \\
\hline
Felbo et al. \cite{felbo2017using} & \checkmark & \checkmark & & & & & & & & & \checkmark & & & \checkmark \\
\hline
Ghosh et al. \cite{ghosh2017magnets} & & & & & & & \checkmark & & & & & & & \\
\hline
Ghosh et al. \cite{ghosh2017role} & \checkmark & \checkmark & & & & \checkmark & & & \checkmark & & & & & \\
\hline
Mukherjee et al. \cite{mukherjee2017sarcasm} & \checkmark & & & & \checkmark & & & & & & & & & \\
\hline
Prasad et al. \cite{prasad2017sentiment} & & & \checkmark & & \checkmark & \checkmark & & & & & & & & \\
\hline
Ghosh et al. \cite{ghosh2018sarcasm} & \checkmark & \checkmark & \checkmark & & \checkmark & & & & \checkmark & & & & & \checkmark \\
\hline
Ilic et al. \cite{ilic2018deep} & \checkmark & & & & & & & & & & & & & \checkmark \\
\hline
Hazarika et al. \cite{hazarika2018cascade} & & & & & & & \checkmark & & & & & & & \checkmark \\
\hline
Tay et al. \cite{tay2018reasoning} & \checkmark & & & & & & & & \checkmark & & & & & \checkmark \\
\hline
Van et al. \cite{van2018semeval} & \checkmark & & & & & & & & & \checkmark & & & & \\
\hline
Wu et al. \cite{wu2018thu_ngn} & & & & & \checkmark & \checkmark & & & \checkmark & & \checkmark & & & \checkmark \\
\hline
Cai et al. \cite{cai2019multi} & \checkmark & & & & & & & \checkmark & & & & \checkmark & & \checkmark \\
\hline
Kumar et al. \cite{kumar2019sarcasm} & & \checkmark & \checkmark & & & & & & & & & \checkmark & & \\
\hline
Majumder et al. \cite{majumder2019sentiment} & & & & & & \checkmark & & & & & & & & \\
\hline
Mehndiratta et al. \cite{mehndiratta2019identification} & \checkmark & & & & & & & & & & \checkmark & \checkmark & \checkmark & \\
\hline
Onan et al. \cite{onan2019topic} & \checkmark & \checkmark & & & & & & & & & \checkmark & \checkmark & \checkmark & \\
\hline
Chakrabarty et al. \cite{chakrabarty2020r} & \checkmark & & & \checkmark & & & \checkmark & & & & & & \checkmark & \\
\hline

\end{longtable}
\end{center}

\section{Methodologies}\label{sec6}
In this section, we discussed and evaluated the approaches and methodologies that have been used in sarcasm detection over the years. 

\subsection{Rule-based Approaches}\label{sec6.1}
Rule-based approaches simply consist of a set of rules that represent sarcastic cues and indicators. Rule based approaches do not need training as it is merely a set of rules based on the data given. Earlier works tend to formulate rule-based or pattern based approaches for sarcasm detection \cite{gonzalez2011identifying,davidov2010semi}. Maynard et al. \cite{maynard2014cares} utilized the hashtag from the provided tweet using a hashtag tokenizer to determine whether the sentiment of the rest of the tweet conflicts with the hashtag or not; if it does, the tweet will be sarcastic. Riloff et al. \cite{riloff2013sarcasm} proposed a rule-based classifier which uses a bootstrapping algorithm to find out a positive verb with a negative situation in a tweet. Bharti et al. \cite{bharti2015parsing} presented two rule-based classifiers for sarcasm detection on a dataset collected from twitter. One of the classifiers tries to capture hyperboles by finding out the interjection and intensifiers occurring together. The second classifier makes use of a parse-based lexicon generation algorithm that builds parse trees of sentences and finds situational expressions that convey emotion. Bharti et al. \cite{bharti2017sarcastic} tried to structure sarcasm and based on various types of sarcasm noticed in twitter posts, the authors proposed six rule-based algorithms for sarcasm detection. They are - Parsing-based algorithm for lexicon generation (PBALN), Tweet Start with Interjection Word (TSIW), Tweet Start with Interjection Word (TSIW), Contradiction between Tweets and the Universal Facts (CTUIFs), Contradiction between Tweet and its Temporal Facts (CTTFs) and Positive Tweet that Contains a Word and its Antonym Pair (PTCAP). These algorithms proved to perform better in terms of precision, recall and F-measure than the state-of-the-art approaches back then. Abulaish et al. \cite{abulaish2018self} used a rule-based technique which uses regular expressions to implement filtering rules, instead of string matching for the first layer of the model which identifies self-around tweets that are candidates for self-deprecating sarcasm. To further classify sarcastic tweets into different categories of sarcasm, Sundararajan et al. \cite{sundararajan2020multi} proposed a rule-based method consisting of a fuzzy rule-based type detection and rough set-based type detection and validation. Rendalkar et al. \cite{rendalkar2018sarcasm} proposed a hybrid approach which combines two approaches, a word based detection that uses SentiNetWord\footnote{\url{https://github.com/aesuli/SentiWordNet}} to get the emotion of the input and an Emoticon based approach to utilize emoticons present in comments to detect sarcasm. Even if there is no training required, as the set of rules are based on the data, the type of sarcastic occurrences are dependent on languages, locations and also different types of social media platforms. The same rules will not be applicable to a new and versatile dataset because the model does not learn anything from the new data. Also, the fact that sarcasm does not always adhere to rules and can have a variety of structures and indicators, makes it difficult to create a comprehensive set of guidelines for the vast amount of data on social media. 

\subsection{Traditional Machine Learning-based Approaches}\label{sec6.2}
We came across many works which utilized machine learning classifiers for sarcasm detection since the earlier years. As seen in numerous works, SVM is one of the most popular classifiers \cite{joshi2015harnessing,davidov2010semi,joshi2016word}. In the work of Gnzalez et al. \cite{gonzalez2011identifying}, we saw SVM being used with SMO (Sequential Minimal Optimization \cite{platt1998sequential}) and logistic regression. Riloff et al. \cite{riloff2013sarcasm} showed that their SVM-based model performed better than the baseline rule-based techniques. SVM even outweighed Neural Networks in sarcasm detection in the work of Jansi et al. \cite{jansi2018extensive}. With their SVM-Word2Vec model, Oraby et al. \cite{oraby2017you} attempted to distinguish rhetorically sarcastic questions, and they discovered that the best outcome came from including post-level scores from the Linguistic Inquiry and Word Count (LIWC) (Pennebaker et al. \cite{pennebaker2001linguistic}) tool in the model. On the other hand, Abercrombie et al. \cite{abercrombie2016putting} and Bamman et al. \cite{bamman2015contextualized} both used logistic regression to find out if the sentence is sarcastic or not. Mukherjee et al. \cite{mukherjee2017sarcasm} showed that in a small dataset, Naive Bayes works better than an unsupervised fuzzy c-means clustering model for sarcasm detection. In machine learning approaches, we noticed a tendency to make use of appropriate features that can be indicators for sarcasm. Mukherjee et al. \cite{mukherjee2017detecting} showed that Naive Bayes also outperformed Maximum Entropy for most of the tweet datasets utilizing content and function words as features. Thakur et al. \cite{thakur2018detecting} utilized Naive Bayes to show that POS tags are not a particularly useful feature in sarcasm detection. Parde et al. \cite{parde2018detecting} used Naive Bayes for their domain-general sarcasm detection both for Twitter and Amazon product reviews. Bharti et al. \cite{bharti2017sarcasm} extracted lexical, hyperbolic, behavioral features and universal facts with their proposed algorithm PBLGA(parsing-based lexical generation algorithm) and showed that Decision Tree performed better than SVM, Naive Bayes and Maximum Entropy for sarcasm detection. SVM with Radial Basis Function (RBF) Kernel outperformed Decision Tree in the experiment by Sreelakshmi et al. \cite{sreelakshmi2018effective}. Prasad et al. \cite{prasad2017sentiment} also compared various machine learning approaches incorporating their emoji and slang dictionary where Gradient Boosting outperformed the other classifiers. Incorporating BERT and GloVe embeddings as features with Logistic Regression worked very well in the work of Khatri et al. \cite{khatri2020sarcasm}. GloVe Embeddings also worked better with Random Forest than SVM and Decision Tree (Eke et al. \cite{eke2020significance}). Random Forest and Ada-Boost based ensemble algorithms are also used in ensembling the features for sarcasm detection (Sundararajan et al. \cite{sundararajan2021textual}). Banerjee et al. \cite{banerjee2020synthetic} tried to remove class imbalance with minority oversampling techniques and proposed that lazy learners like KNN are not suitable for sarcasm detection when minority oversampling is used. To avoid the strenuous feature engineering process, Di et al. \cite{di2019effectiveness} exploited a Distributional Semantics approach using latent semantic analysis (LSA) (Landauer et al. \cite{landauer1998introduction}) and tested out with various machine learning classifiers (SVM, Logistic Regression, Random Forests, and Gradient boosting). The problem with traditional machine learning models is the need to construct a set of hand-crafted features to make the patterns of the data visible to the learning algorithm and reduce complexity. We need to be knowledgeable about that particular domain and data patterns for the hard core feature extraction where the features are the most suitable for sarcasm detection. 

\subsection{Deep Learning-based Approaches}\label{sec6.3}
With the development of more deep learning models, we noticed a trend of using deep learning models or hybrid models (mix of deep and traditional machine learning models) rather than only traditional machine learning models for sarcasm detection. Deep learning models became popular because of their capacity to extract features automatically, allowing us to avoid collecting hand-crafted features. Poria et al. \cite{poria2016deeper} were the first to use a pre-trained CNN to automatically extract sentiment, emotion and personality features and feed them to a SVM for the final classification. We saw a lot of different deep learning models being deployed together such as in the experiment by Ghosh et al. \cite{ghosh2016fracking}, to exploit the semantic modeling power the authors presented a CNN model whose output is fed into an LSTM layer before being fed into a DNN layer. This model seemed to outperform the work of Riloff et al. \cite{riloff2013sarcasm} and Davidov et al. \cite{davidov2010semi} with an F1-score of 0.92. Deep learning models have been seen to easily outperform the traditional machine learning models. Porwal et al. \cite{porwal2018sarcasm} and Salim et al. \cite{salim2020deep} both created a NN model with RNN and LSTM that extracts features automatically. Guo et al. \cite{guofinding} showed that, for their Reddit dataset, LSTM outperformed the baseline Bag-of-Words, Naive Bayes and even the traditional ‘vanilla’ neural network as LSTM allows the network to more effectively learn sequential data without overfitting, utilizing the contextual data for sarcasm detection. Mehndiratta et al. \cite{mehndiratta2019identification} showed that for sarcasm detection, their single LSTM model performed even better than their hybrid LSTM-CNN model. \\

Ghosh et al. \cite{ghosh2017role} showed that in the case of detecting sarcasm in conversations within social media, adding conversational context as one of the inputs to an LSTM model performed better than using LSTM without conversational context. Ghosh et al. \cite{ghosh2018sarcasm} used the previous sentence of the sarcastic utterance as context and showed that a multiple-LSTM architecture with sentence level attention gave better performance than a single LSTM architecture. To model the semantic relationship between the candidate text and its context, Diao et al. \cite{diao2020multi} built an end-to-end Multi-dimension Question Answering model based on Bi-LSTM and attention mechanism using multi-granularity representations which outperformed the state-of-the-art machine learning model by a cosiderable margin in case of F1 score. Ren et al. \cite{ren2018context} experimented with two types of context, history based context (views and opinions towards some events and people) and conversational context to see which type better helps in sarcasm detection. They investigated with two types of context augmented CNNs named CANN-Key and CANN-ALL where CANN-KEY integrates key contextual information and CANN-ALL integrates all contextual information. They found that for conversation-based contexts, CANN-ALL has more capability in capturing minor hints of sarcasm but history based contexts achieved better results as the number of the conversation-based tweets are relatively small. \\

As sarcastic nature and form of expression vary from person to person, Hazarika et al. \cite{hazarika2018cascade} proposed a model called CASCADE (a ContextuAl SarCasm DEtector) which jointly utilizes context information from discussion forums and user embeddings that encode stylometric and personality features of the users with a CNN-based textual model. Experiments showed that adding the personality features along with the context further improved the model’s performance. Kolchinski et al. \cite{kolchinski2018representing} also tried to model author embeddings with two methods - a simple Bayesian method that captures solely an author’s raw propensity for sarcasm, and a dense embedding method that allows for intricate interactions between author and text. Then these author embeddings extend a baseline bidirectional RNN with GRU cells (BiGRU) which models all the user comments. The method slightly under-performed than CASCADE on the full SARC\footnote{\url{https://nlp.cs.princeton.edu/SARC/}} dataset but outperformed CASCADE on the posts of the sub-reddit named r/politics of the same dataset. On the contrary, Misra et al. \cite{misra2019sarcasm} proposed that without common sense knowledge and current events’ information, the model does not understand properly and just picks up discriminative lexical cues. So instead, they removed the user-embeddings and focused on using current events and common sense knowledge with their LSTM-CNN module which improved upon the baseline models by $\sim$5\%.The need of common sense knowledge in sarcasm detection is further discussed in section \ref{sec8.4}. The problem with these sequence to sequence models is that we can see a decline in performance when dealing with longer sentences as it compresses all the information of an input source sentence into a fixed-length vector which may lose relevant information. \\

To solve the long range dependency problem, in the later works, we saw the use of attention based deep learning models as the relevance of each word in a sarcastic statement is learned by the attention layer, which then gives each word a different weight. Kumar et al. \cite{kumar2020sarcasm} presented a Multi-Head self-Attention based Bidirectional LSTM (MHA-BiLSTM) modeled with manually designed auxiliary features which outperforms a feature-rich SVM model (semantic, sentiment and punctuation features). This SVM previously performed better than only a BiLSTM model but fell behind the MHA-BiLSTM model by a significant margin of 4.45\% and 7.88\% on balanced dataset and imbalanced dataset, respectively. To increase interpretability of the model along with its high performance, Liu et al. \cite{liu2021dual} used GRU with Multi-Head self-Attention which provides crucial cues for sarcasm. Kumar et al. \cite{kumar2019sarcasm} implemented another attention based hybrid model that combines soft-attention based BiLSTM with punctuation based auxiliary pragmatic features to a deep convolution network for enhanced performance. Along with the automated extracted features from their Hybrid attention based LSTM model, Pandey et al. \cite{pandey2021hybrid} combined 16 hand crafted features to explore the role of different linguistic features on sarcasm detection. Adding attention mechanism improved the problem of long range dependencies in sequence models, but dealing with it is still challenging and also the sequential nature of the models prevents parallel processing. Transformer-based models solve these problems as we will see in section \ref{sec6.4}. \\
 
GRU and LSTM parse words one at a time which hampers modeling contrast, incongruity and long-range dependencies of multiple sentences. To tackle this, Tay et al. \cite{tay2018reasoning} was the first to propose a Multi-dimensional Intra-Attention Recurrent Network (MIARN) which leverages intra-sentence relationships based on the intuition of compositional learning. MIARN outperformed models like NBOW, CNN, LSTM, ATT-LSTM (Attention-based LSTM), GRNN (Gated-RNN), CNN-LSTM-DNN across all the six datasets that were used. Akula et al. \cite{akula2021explainable} used MIARN as encoders for their Dual-Channel Network (DC-Net) to model both literal and deep meanings of sentiments to recognize the sentiment conflict of the input texts. Pan et al. \cite{pan2020modeling} introduced snippet-level self-attention to model the incongruity between sentence snippets where the model consists of a convolution module (CNN), an importance weighting module, and a self-attention module which worked better on twitter datasets rather than long text datasets. 

\subsubsection{Multi Task Learning}\label{sec6.3.1}
Multi task learning is a novel learning scheme which uses a single neural network to perform more than one classification task at the same time. In our case, we discovered a few works that used multitask learning to detect sarcasm. Majumder et al. \cite{majumder2019sentiment} used multi task learning for both sentiment classification and sarcasm detection. They achieved sentence representations using GRU with attention mechanism and Glove word-embeddings for word representations. They employed two distinct softmax layers for classification for the two tasks. Adding NTN (neural tensor network) to the multitask classifier further improved results, giving a better performance for sarcasm detection. Similarly, Savini et al. \cite{savini2020multi} made use of multi-task learning to inform the primary task of sarcasm detection by using sentiment classification as an auxiliary task. Both tasks use the same BiLSTM model and ELMo\footnote{\url{https://allenai.org/allennlp/software/elmo}} and FastText embeddings but have different Multi-layer Perceptron and no user embeddings. This model outperformed many systems that model user embeddings (CNN-SVM, CUE-CNN) but its F1-score is only 0.7\% lower than CASCADE. 

\subsubsection{Ensemble Learning}\label{sec6.3.2}
Few works utilized ensemble learning techniques rather than using one specific classifier for sarcasm detection. Jain et al. \cite{jain2017sarcasm} used two different ensemble learning methods, Random Forest and Weighted Ensemble as their sarcasm detection classifier. The weighted ensemble model uses Naive Bayes, Linear Regression, and Random Forest as its component classifiers. According to them, ensemble-based approaches are more effective in terms of recall and precision, and their effectiveness is largely dependent on the effectiveness of the individual classifiers. A Deep Ensemble Soft Classifier (DESC) comprising of three deep models — a BiLSTM, an AttentionLSTM, and a Dense NN was proposed by Potamias et al. \cite{potamias2019robust}. DESC outperformed all the models published in SemEval-2015 Sentiment Analysis task. Gupta et al. \cite{gupta2020statistical} used a voting classifier which uses majority voting to choose the best result among results given by the several machine learning classifiers. Lemmens et al. \cite{lemmens2020sarcasm} proposed an ensemble method that is trained using both additional features and the predicted sarcasm probabilities of four component models. The component models consist of an LSTM with hashtag and emoji representations; a CNN-LSTM with casing, stop word, punctuation, and sentiment representations; an MLP (Multi-layer Perceptron) based on InferSent embeddings\footnote{\url{https://github.com/facebookresearch/InferSent}}; and an SVM trained on stylometric and emotion-based features. 

\subsection{Transformer based Approaches}\label{sec6.4}
Transformer models make use of attention blocks which take account of short and long range dependencies with the same likelihood, overcoming the shortcomings of the previous architectures\footnote{\url{https://towardsdatascience.com/all-you-need-to-know-about-attention-and-transformers-in-depth-understanding-part-1-552f0b41d021}}. Most of the recent architectures frequently tend to use Transformer models namely BERT or some variant of BERT along with RoBERTa (Iu et al. \cite{liu2019roberta}), ALBERT (Lan et al. \cite{lan2019albert}) etc. Srivastava et al. \cite{srivastava2020novel} used a hierarchical BERT-based model, which consists of a context-summarization layer, a context-encoder layer, a CNN-Layer, and finally a fully-connected layer, to process the input contexts for a given response. Gregory et al. \cite{gregory2020transformer} explored new methods with various transformers to classify sarcasm in twitter posts. They saw that, among the individual models, BERT performed better. But their best performing model was an Ensemble model, which ensembled the results of five separate pre-trained transformer models - BERT, RoBERTa, XLNet, RoBERTa-large, and ALBERT. They assumed that the ensemble model allows more information in each of the embeddings as opposed to a single form of embedding. Kalaivani et al. \cite{kalaivani2020sarcasm} compared BERT’s performance with the traditional machine learning (SVM, Naive Bayes, Logistic Regression) and deep learning models (RNN-LSTM) and showed that BERT performs the best among all the models as BERT works well with continuous conversation dialogues. To reduce the overhead of data-preprocessing, Potamias et al. \cite{potamias2020transformer} was the first to implement an end-to-end model that used an unsupervised pre-trained transformer method in figurative language. The pre-trained RoBERTa was combined with a RCNN (Recurrent Convolutional Neural Network) to capture different forms of contextual information and it contains no handcrafted engineered features or lexicon dictionaries. Javdan et al. \cite{javdan2020applying} proposed to use a combination of aspect-based sentiment analysis and BERT, in order to detect sarcasm in Twitter and Reddit. The individual BERT model performed best in Reddit, whereas LCF-BERT, an aspect-based sentiment classification method, introduced in the work of Zeng et al. \cite{zeng2019lcf} performed best on the Twitter dataset. As aspect-based methods try to learn the input data more interactively, taking them as two separate sections works better with less complex data like Twitter posts. Parameswaran et al. \cite{parameswaran2021bert} fine-tuned two publicly available BERT models - TD-BERT (Gao et al. \cite{gao2019target}) which incorporates the potential target information in its classification input, and BERT-AEN (Song et al. \cite{song2019attentional}) which uses an attention encoder network to model the semantic interaction between the given sentence and the potential target, to show that BERT models outperform the current state-of-the-art models for sarcasm target detection. BERT-AEN's multiple attention mechanism was anticipated to work better than TD-BERT, but surprisingly TD-BERT even performed better than BERT-AEN along with the other baseline models. Simply incorporating the target’s position helped TD-BERT understand the context better which may have led to the performance boost. Kumar et al. \cite{kumar2021adversarial} came up with a new model for sarcasm detection, Adversarial and Auxiliary Features-Aware BERT (AAFAB) which tries to encode the semantic meaning of a sentence and combines with high quality manually extracted auxiliary features. Along with BERT word embedding, BERT encoding and feature concatenation, another step called adversarial training was conducted by adding perturbations to the input word embedding to improve parameter generalization. AAFAB outperformed several deep learning-based baseline models on both balanced and imbalanced datasets. Lou et al. \cite{lou2021affective} is the first to propose an Affective Dependency Graph Convolutional Network (ADGCN) framework for sarcasm detection. This framework constructs an affective graph and a syntax-aware dependency graph for each of the sentences based on affective commonsense knowledge and dependency trees. For sarcasm detection, it uses BERT to learn the vector representations of the context and multi-layer GCNs to leverage the affective dependencies of the context. To collect less noisy sarcastic data using conversation cues, Shmueli et al. \cite{shmueli2020reactive} proposed a novel data collection technique called reactive supervision. They created a new large dataset with extra features and fine-tuned labels, named SPIRS\footnote{\url{https://paperswithcode.com/dataset/spirs}}. Along with sarcasm detection, they enabled a new task of sarcasm perspective classification where the model tries to find out if the sarcasm is intended or not. In the evaluation, pre-trained BERT outperformed other deep learning methods. With parallel processing, computation is faster for a transformer model but the model has an inability to process hierarchical inputs where it cannot leverage the highest level representations of the input sequence from the past to compute the current representation as the computation is done in parallel for every step (Fan et al. \cite{fan2020addressing}). \\

Table \ref{tab3} shows a comparative analysis between the performances of various sarcasm detection systems. This should be noted, that the entries are not comparable to each other as the experiments were not done with the same datasets and conditions.


\begin{center}
\begin{longtable}{|p{0.12\textwidth}|p{0.09\textwidth}|p{0.19\textwidth}|p{0.09\textwidth}|p{0.08\textwidth}|p{0.08\textwidth}|p{0.08\textwidth}|}
\caption{Performance summary of various approaches used in sarcasm detection}
\label{tab3} \\
\hline
&\textbf{Data}&\textbf{Architecture}&\multicolumn{4}{|c|}{\textbf{Performance}}\\
\hline
& & & \begin{sideways}\textbf{Accuracy}\end{sideways} & \begin{sideways}\textbf{F1-Score}\end{sideways} & \begin{sideways}\textbf{Precision}\end{sideways} & \begin{sideways}\textbf{Recall}\end{sideways} \\
\hline
Davidov et al.\cite{davidov2010semi} & Tweets & SASI (Semi-supervised Algorithm for Sarcasm Identification) & 0.896 & 0.545 & 0.727 & 0.436\\
\hline
Gupta et al. \cite{gupta2017crystalnest} & Tweets & CrystalNet & & 0.60 & 0.52 & 0.70\\
\hline
Bharti et al. \cite{bharti2017sarcasm} & Tweets & PBLGA with SVM & & 0.67 & 0.67 & 0.68 \\
\hline
Mukherjee et al. \cite{mukherjee2017detecting} & Tweets & Naive Bayes & 0.73 & & & \\
\hline
Jain et al. \cite{jain2017sarcasm} & Tweets & Weighted Ensemble & 0.853 & & 0.831 & 0.298 \\
\hline
Poria et al. \cite{poria2016deeper} & Tweets & CNN-SVM &  & 0.9771 &  & \\
\hline
Ghosh et al. \cite{ghosh2016fracking} & Tweets & CNN-LSTM-DNN &  & 0.901 & 0.894 & 0.912 \\
\hline
Zhang et al. \cite{zhang2016tweet} & Tweets & GRNN & 0.9074  & 0.9074 &  &  \\
\hline
Oraby et al. \cite{oraby2017you} & Tweets & SVM + W2V + LIWC &  & 0.83 & 0.80  & 0.86 \\
\hline
Hazarika et al. \cite{hazarika2018cascade} & Reddit posts & CASCADE & 0.79 & 0.86 & & \\
\hline
Ren et al. \cite{ren2018context} & Tweets & CANN-KEY &  & 0.6328 & &\\
\hline
& & CANN-ALL & & 0.6205 & & \\
\hline
Tay et al. \cite{tay2018reasoning} & Tweets, Reddit posts & MIARN & Twitter: 0.8647 & 0.86 & 0.8613 & 0.8579 \\
\hline
& & & Reddit: 0.6091 & 0.6922 & 0.6935 & 0.7005\\
\hline
Ghosh et al. \cite{ghosh2018sarcasm} & Reddit posts & multiple-LSTM & 0.7458 & 0.7607 & & 0.7762 \\
\hline
Diao et al. \cite{diao2020multi} & Internet arguments & MQA (Multi-dimension Question Answering model) & & 0.762 & 0.701 & 0.835 \\
\hline
Kumar et al. \cite{kumar2020sarcasm} & Reddit posts & MHA-BiLSTM & & 0.7748 & 0.7263 & 0.8303 \\
\hline
Kumar et al. \cite{kumar2019sarcasm} & Tweets & sAtt-BiLSTM convNet & 0.9371 & & & \\
\hline
Majumder et al. \cite{majumder2019sentiment} & Text snippets & Multi task learning with fusion and shared attention & & 0.866 & 0.9101 & 0.9074\\
\hline
Potamias et al. \cite{potamias2019robust} & reviews of laptops and restaurants & DESC (Deep Ensemble Soft Classifier) & 0.74 & 0.73 & 0.73 & 0.73 \\
\hline
Srivastava et al. \cite{srivastava2020novel} & Tweets, Reddit posts & BERT + BiLSTM + CNN & Twitter: 0.74 & & &\\
\hline
& & & Reddit: 0.639 & & &\\
\hline
Gregory et al. \cite{gregory2020transformer} & Tweets, Reddit posts & Transformer ensemble (BERT, RoBERTa, XLNet, RoBERTa-large, and ALBERT) &  & 0.756& 0.758 & 0.767\\
\hline
Potamias et al. \cite{potamias2020transformer} & Tweets, Reddit politics & RCNN-RoBERTa & Twitter: 0.91 & 0.90 & 0.90 & 0.90\\
\hline
& & & Reddit: 0.79 & 0.78 & 0.78 & 0.78\\
\hline
Javdan et al. \cite{javdan2020applying} & Tweets & LCF-BERT & & 0.73 & & \\
\hline
& Reddit posts & BERT-base-cased &  & 0.734 & & \\
\hline
Lee et al. \cite{lee2020augmenting}  & Tweets, Reddit posts & BERT + BiLSTM + NeXtVLAD & Twitter & 0.8977 & 0.8747 & 0.9219\\
\hline
& & & Reddit & 0.7513 & 0.6938 & 0.8187\\
\hline
Baruah et al. \cite{baruah2020context} & Tweets, Reddit posts & BERT-large-uncased & Twitter & 0.743 & 0.744 & 0.748\\
\hline
& & & Reddit & 0.658 & 0.658 & 0.658\\
\hline
Avvaru et al. \cite{avvaru2020detecting} & Tweets, Reddit posts & BERT & Twitter & 0.752 & &\\
\hline
& & & Reddit & 0.621 & & \\
\hline
Jaiswal et al. \cite{jaiswal2020neural} & Tweets, Reddit posts & Ensemble of several combinations of RoBERTa-large & & 0.790 & 0.790 & 0.792\\
\hline
Shmueli et al. \cite{shmueli2020reactive} & Tweets & BERT & 0.703 & 0.699 & 0.70 0.7741 &\\
\hline
Dadu et al. \cite{dadu2020sarcasm} & Tweets, Reddit posts & RoBERTa-large & Twitter & 0.772 & 0.772 & 0.772\\
\hline
& & & Reddit & 0.716 & 0.716 & 0.718\\
\hline
Kalaivani et al. \cite{kalaivani2020sarcasm} & Tweets, Reddit posts & BERT & Twitter & 0.722 & 0.722 & 0.722\\
\hline
& & & Reddit & 0.679 & 0.679 & 0.679\\
\hline
Naseem et al. \cite{naseem2020towards} & Tweets & T-DICE + BiLSTM + ALBERT & 0.93 & 0.93 & &\\
\hline
Dong et al. \cite{dong2020transformer} & Tweets, Reddit posts & context-aware RoBERTa-large & Twitter & 0.783 & 0.784 & 0.789\\
\hline
& & & Reddit & 0.744 & 0.745 & 0.749\\
\hline
Kumar et al. \cite{kumar2020transformers} & Tweets, Reddit posts & context-aware RoBERTa-large & Twitter & 0.772 & 0.773 & 0.774\\
\hline
& & & Reddit & 0.691 & 0.693 & 0.699\\
\hline
Kumar et al. \cite{kumar2021adversarial} & Tweets & AAFAB (Adversarial and Auxiliary Features-Aware BERT) & & 0.7997 & 0.8101 & 0.7896\\
\hline
Lou et al. \cite{lou2021affective} & Tweets, Reddit posts & ADGCN-BERT (Affective Dependency Graph Convolutional Network) & Twitter: 0.9031 & 0.8954 & & \\
\hline
& & & Reddit: 0.8077 & 0.8077 & & \\
\hline

\end{longtable}
\end{center}

\subsection{Shared Task}\label{sec6.5}
Shared Tasks are organized to tackle a specific problem where the organizers provide the problem set and necessary datasets and research teams try to solve the problem. For the SemEval-2017 sentiment analysis task (Task 4), Gupta et al. \cite{gupta2017crystalnest} incorporated a sarcasm detection mechanism using a linear SVM classifier with their sentiment analysis model, Crystalnet and saw a performance boost proving that proper detection of sarcasm is indeed necessary for understanding the genuine sentiment of a text. At the 2nd Workshop on the Figurative Language Processing 2020\footnote{\url{https://competitions.codalab.org/competitions/22247}}, a shared task on sarcasm detection was presented. As previously mentioned, in section \ref{sec5.1.7}, two datasets on Twitter and Reddit were given to investigate if conversational context was important to detect sarcasm or not (Ghosh et al. \cite{ghosh2020report}). The baseline scores for both the datasets were given by the organizers. The majority of the solution approaches presented some version of Transformer based architectures, specially BERT and RoBERTa, indicating that the trend of sarcasm classification is shifting towards applying pre-trained language models. We noticed that few systems attempted to use varied lengths of context (3,5,7 sentences) as input along with response to explore how context affected sarcastic sentence recognition. Baruah et al. \cite{baruah2020context} showed that for the Reddit dataset, the BERT classifier achieved their individual highest F-score using only response as input and zero context utterances, however in the Twitter dataset, it achieved the highest score using response and the last utterance as context. After incorporating the previous utterance with the response as input and using RoBERTa-large, ALBERT, and BERT for classification, Dadu et al. \cite{dadu2020sarcasm} and Dong et al. \cite{dong2020transformer} both witnessed an increase in performance in both the datasets where RoBERTa-large gave the highest improvement. Jaiswal et al. \cite{jaiswal2020neural} implemented the second best performing model for the given task. In their case, choosing the latest three utterances gave the best result using both BERT and RoBERTa. For the classification model, they used an approach known as `majority voting' in which several models predict the outcome and the label is determined by the majority's output. Avvaru et al. \cite{avvaru2020detecting} showed a promising outcome of using seven latest context utterances along with the response as input. Lee et al. \cite{lee2020augmenting} presented the best solution for the shared task, an architecture including BERT and pooling layers decomposed of BiLSTM and NeXtVLAD \cite{lin2018nextvlad}. But what improved the result was the data augmentation technique called CRA (Contextual Response Augmentation), which enlarged the given dataset with more samples. CRA makes use of easily accessible conversational context of unlabeled dialogue threads from Reddit and Twitter and encodes each response in the labeled training set using the BERT trained on natural inference tasks, introduced in the work of Reimers et al. \cite{reimers2019sentence}.

\subsection{Beyond Sarcasm Detection}\label{sec6.6}
As observed in our investigation, the field of study has expanded beyond sarcasm detection. Dubey et al. \cite{dubey2019numbers} is the first to propose various rule-based, machine learning and deep learning models to detect sarcasm using the numerical portions of the input. Their deep learning model, CNN-FF with an attention model outperformed other models in the experiment. Patro et al. \cite{patro2019deep} used deep learning to find out the target of sarcasm where the word embeddings are passed on to a bidirectional LSTM (Bi-LSTM) layer or a target dependent LSTM (TD-LSTM) layer. A few other works also attempted to identify the intended target of sarcasm. Joshi et al. \cite{joshi2016automatic} was the first one to work on this and they used SVMperf with two rule based extractors as their classifier. There has been a few attempts to generate computational sarcasm in recent years. Sarcasm generation is a challenging task since the generated utterance should possess a number of properties of sarcastic texts.  The first attempt at automatic sarcasm generation was made by Mishra et al. \cite{mishra2019modular}, whose system relies on the theory of context incongruity and anticipates input with a negative sentiment. The model made use of LSTM and was tested against SarcasmBot \cite{joshi2015sarcasmbot}, UNMT \cite{artetxe2017unsupervised}, Monoses \cite{artetxe2018unsupervised}, ST \cite{shen2017style} and FLIP. With the help of Transformer models, Chakrabarty et al. \cite{chakrabarty2020r} proposed an unsupervised sarcasm generation technique where a non-sarcastic sentence is given as input. Their method tries to incorporate reversal of valence and semantic incongruity with the context, the two main characteristics of sarcasm, into the sentences. RoBERTa-large is used to incorporate semantic incongruity. They found that the proposed system generates 34\% better sarcastic sentences than human judges. Dubey et al. \cite{dubey2019deep} tried to generate a non-sarcastic interpretation of the sarcastic counterpart for the input text with their rule-based, deep learning-based and machine learning-based architectures. They observed that their statistical machine translation-based approach which used Moses (an Open Source Toolkit for Statistical Machine Translation) \cite{koehn2007open}, outperformed other approaches on the first dataset. 

\section{Trends}\label{sec7}
\begin{figure*}[htbp]
\centering

\tikzstyle{descript} = [text = black,align=center, minimum height=1cm, align=center, outer sep=0pt,font = \footnotesize]
\tikzstyle{activity} =[align=center,outer sep=1pt]

\caption{Trends in sarcasm detection throughout the years\\}
\label{figTrends}

\begin{tikzpicture}[very thick, black]
\small

\coordinate (O) at (0,0); 
\coordinate (F) at (11.25,0); 

\coordinate (E1) at (1.75,0); 
\coordinate (E2) at (2.35,0); 
\coordinate (E3) at (5.25,0); 
\coordinate (E4) at (6.125,0); 
\coordinate (E5) at (9,0); 
\coordinate (E6) at (10,0); 
	
\draw[-,thick,color=black] ($(E1)+(0,0.1)$) -- ($(E1)+(0,1)$) node [above=90pt,align=center,black] {\textbf{Foundational} \\ \textbf{Research}};
\draw[-,thick,color=black] ($(E1)+(0,0.1)$) -- ($(E1)+(0,1)$) node
[above=-10pt,align=left,black] {
First paper on \\sarcasm detection\\ \textit{Tepperman et al.\cite{tepperman2006yeah}}\\ \\
Sarcasm detection using \\Lexical features\\ \textit{Kreuz et al.\cite{kreuz2007lexical}}\\
};

\draw[-,thick,color=black] ($(E2)-(0,0.1)$) -- ($(E2)-(0,1)$) node [below=0pt,align=center,black] {\textbf{Pattern \& Rule} \\ \textbf{-based Methods}};
\draw[-,thick,color=black] ($(E2)-(0,0.1)$) -- ($(E2)-(0,1)$) node
[below=25pt,align=left,black] {
Semi-supervised \\approach in Twitter\\ \textit{Davidov et al.\cite{davidov2010semi}}\\ \\
Impotance of Linguistic \\ \& Fragmatic Features\\\textit{González-Ibáñez et al.\cite{gonzalez2011identifying}}\\
};

\draw[-,thick,color=black] ($(E3)+(0,0.1)$) -- ($(E3)+(0,1)$) node [above=90pt,align=center,black] {\textbf{Distant Supervision} \\ \textbf{with Hashtags}};
\draw[-,thick,color=black] ($(E3)+(0,0.1)$) -- ($(E3)+(0,1)$) node
[above=-10pt,align=left,black] {
Hashtag-based \\Annotation\\ \textit{ Liebrecht et al.\cite{liebrecht2013perfect}}\\ \\
Using hashtags \\as features\\ \textit{Maynard et al.\cite{maynard2014cares}}\\
};

\draw[-,thick,color=black] ($(E4)-(0,0.1)$) -- ($(E4)-(0,1)$) node [below=0pt,align=center,black] {\textbf{Integrating Context}};
\draw[-,thick,color=black] ($(E4)-(0,0.1)$) -- ($(E4)-(0,1)$) node
[below=25pt,align=left,black] {
Necessity of context in\\ sarcasm detection\\ \textit{Wallace et al.\cite{wallace2014humans}}\\ \\
Using conversation, \\author and \\audience context\\ \textit{Bamman and Smith al.\cite{bamman2015contextualized}}\\
};

\draw[-,thick,color=black] ($(E5)+(0,0.1)$) -- ($(E5)+(0,1)$) node [above=90pt,align=center,black] {\textbf{Deep Learning}};
\draw[-,thick,color=black] ($(E5)+(0,0.1)$) -- ($(E5)+(0,1)$) node
[above =-10pt,align=left,black] {
Using Deep Convolution \\Neural Network\\ \textit{Poria et al.\cite{poria2016deeper}}\\ \\
Neural network semantic \\ model with CNN,LSTM \& \\DNN \textit{Ghosh et al.\cite{ghosh2016fracking}}\\
};

\draw[-,thick,color=black] ($(E6)-(0,0.1)$) -- ($(E6)-(0,1)$) node [below=0pt,align=center,black] {\textbf{Transformers}};
\draw[-,thick,color=black] ($(E6)-(0,0.1)$) -- ($(E6)-(0,1)$) node
[below=25pt,align=left,black] {
Using unsupervised\\ pre-trained transformers \\\textit{Potamias et al. \cite{potamias2020transformer}}\\ \\ \\
Transformers in \\ multimodal environment \\\textit{Wang et al.\cite{wang2020building}}\\
};

\draw[->] (O) -- (F);

\end{tikzpicture}

\end{figure*}

In this section, we explored different trends in the prior works of computational sarcasm detection. The trends are displayed in a summarized form in figure \ref{figTrends}. \\

Six significant milestones have been reached in sarcasm detection, as shown in figure \ref{figTrends} - 1. Foundational Research, 2. Pattern \& Rule-based Methods, 3. Distant Supervision with Hashtags, 4. Integrating Context, 5. Deep Learning and 6. Transformers. \\

The first known work on sarcasm detection was by Tepperman et al. \cite{tepperman2006yeah}. Following this, many studies experimented with supervised and semi-supervised approaches that focused on discovering patterns and feeding these patterns as features to a statistical or rule-based classifier. When Twitter became a viable source of data, hashtag-based remote monitoring became widespread. After this, a trend of utilizing contextual information such as author, audience, conversation, visual data and so on emerged. Recently, most of the studies have shown interest in deep learning and transformer-based techniques. \\

With this section, we aim to give readers a comprehensive idea of the current practices in sarcasm detection in social networking sites. For this reason, our study will be primarily focusing on the more recent trends - \ref{sec7.1} Integrating Context, \ref{sec7.2} Deep Learning, and \ref{sec7.3} Transformers. 

\subsection{Integrating Context}\label{sec7.1}
In recent years, incorporating the context of the text that we intend to predict on has grown in popularity. From here on till the end of the section, we will be using ‘target text’ for indicating the text that needs to be classified and ‘context’ would mean any information other than the target text. Contextual information may include conversational context, author context, visual context, target context or cognitive features \cite{ghosh2018sarcasm}. Wallace et al. \cite{wallace2014humans} was the first to study and advance the significance of context in detecting sarcastic texts. While using the Bag-of-Words(BoW) method in their Reddit irony corpus\footnote{\url{https://github.com/bwallace/ACL-2014-irony}}, they noticed that the texts that the machine learning algorithm misclassified, were also the ones for which their human annotators frequently requested additional context. Based on such reasoning, they argued that since the human annotators need context, then contextual information should also be provided to the machine learning algorithms. Many future studies were encouraged by the findings of Wallace et al. \cite{wallace2014humans} to include context in their sarcasm detection algorithms \cite{joshi2016would,kolchinski2018representing,plepi2021perceived}. We came across several architectures that took context into account. So far, we have encountered 3 types of contexts - topical context, authorial context, and conversational context. 

\subsubsection{Topical Context}\label{sec7.1.1}
Topical context, as the name suggests, typically refers to the topic of the target text. Wang et al. \cite{wang2015twitter} considered sarcasm detection as a sequential classification task and made use of such topic-based context in their study. They generated a Twitter dataset by deriving the whole tweet sequence, including several tweets with the same hashtag preceding the target tweet for the topic-based context. Joshi et al. \cite{joshi2016would} proposed a topic model for identifying sarcasm-prevalent topics and topic level sentiment using a Twitter dataset. They discovered that topics like ‘work’, ‘gun laws’, ‘weather’ were more prevalent to sarcasm at the time of their study.

\subsubsection{Authorial Context}\label{sec7.1.2}
Authorial context refers to the traces left by the author of the target text. Tay et al. \cite{tay2018reasoning} correlated sarcasm with the discrepancy between the expressed emotion and the author's circumstance (context). Bamman et al. \cite{bamman2015contextualized} extracted extra-linguistic data like the author historical salient terms, author historical topics, profile information, author historical sentiment and profile unigrams for authorial context. These features, combined with the context from intended (or perceived) audience and the conversation between them, gave them a significant increase in accuracy compared to the previous studies. This set a new benchmark for sarcasm detection in social platforms. Mukherjee et al. \cite{mukherjee2017sarcasm} attempted to capture the writing style of the author and discovered authorial traits such as function words and part of speech n-grams, particularly function words, to be critical for identifying sarcasm. Ghosh et al. \cite{ghosh2017magnets} proposed to take the author's mood into consideration. Using a deep neural network architecture, they modeled the author at the time of utterance creation using mood indicators taken from the most recent tweets. The context was modeled using attributes derived from the response utterance's proximate cause.  

\subsubsection{Conversational Context}\label{sec7.1.3}
By conversational context, we refer to the conversation or discussion between the author and the audience of the target text. To capture the conversational context between the target tweet and the tweet to which it is responding, Bamman et al. \cite{bamman2015contextualized} extracted binary indicators of pairwise Brown features between the original and response messages. Wang et al. \cite{wang2015twitter} collected the entire tweet thread, including tweets preceding the target tweet that represent the conversation with other users. Ghosh et al. \cite{ghosh2017role} followed both of these methods for building their conversational context that contained 25,991 instances. Ren et al. \cite{ren2018context} made use of the same dataset created by Wang et al. \cite{wang2015twitter}. Since Wang et al. \cite{wang2015twitter} used SVM\textsuperscript{multiclass} \cite{altun2003hidden} and SVM\textsuperscript{hmm} \cite{vanzo2014context} in their study, Ren et al. \cite{ren2018context} opted for a neural network architecture like CNN to evaluate the contribution of context. Ghosh et al. \cite{ghosh2018sarcasm}, on the other hand, chose LSTM and conditional LSTM networks for verifying the effectiveness of conversational context in identifying sarcasm. They considered the previous turn, the succeeding turn, or both as conversational context. Besides these studies, conversational context has widely been used in Transformer-based approaches. More on this is discussed in the Transformers subsection (section \ref{sec7.3}).

\subsection{Deep Learning}\label{sec7.2}
Using deep learning models for Natural Language Processing research has been around for about a decade, the first one dating back to 2011 by Collobert et al. \cite{collobert2011natural}. However, the first work which utilized deep learning in the domain of sarcasm detection was performed in 2016 by Poria et al. \cite{poria2016deeper}. Using CNN, they trained 3 models, namely: sentiment, emotion and personality, each with their corresponding dataset and the features extracted from these pre-trained models were then fed to a SVM which ultimately classifies the text. Their architecture outperformed the previous state-of-the-art methods. In addition to this, their framework also demonstrated the significance of sentiment shifting, emotion and personality traits for sarcasm detection. Amir et al. \cite{amir2016modelling} also proposed a CNN based model that made use of prior utterances to learn and exploit user embeddings. Instead of considering sarcasm detection and sentiment analysis as separate tasks, Majumder et al. \cite{majumder2019sentiment} developed an architecture that showed and utilized the correlation between the two. With the addition of NTN fusion, their multitask framework improved the performance of sarcasm detection while the inclusion of an attention network shared by both tasks increased the performance of sentiment classification. As deep learning was introduced, different types of data aside from the usual textual form was fed to the models for training. Two other such forms of data, visual and numerical, are discussed in detail in the following subsections.

\subsubsection{Visual Data}\label{sec7.2.1}
It may not always be possible to determine sarcasm from language alone without knowing the underlying meaning of the text. If we consider a tweet - “Will be at office in no time.” – followed by an image of being stuck in traffic jam, then without understanding the context of the image, this will not be classified as a sarcastic tweet. For exploring such scenarios, many studies started incorporating image data to detect sarcasm. The impact of visual content for sarcasm detection in social media was first empirically investigated by Schianella et al. \cite{schifanella2016detecting}. They used deep learning to combine a deep network-based depiction of the image data by taking unigrams as text inputs. The Hierarchical Fusion Model of Cai et al. \cite{cai2019multi} is another notable work that utilized image features in their sarcasm detection architecture. They used ResNet to obtain regional vectors of images and to predict 5 attributes of each image. Bi-LSTM was used to obtain text vectors. The feature vectors were then reconstructed with the help of the raw vectors and the guidance vectors. After that, these refined vectors were further fused together into one vector. Their deep learning architecture created a benchmark for multi-modal sarcasm detection. 

\subsubsection{Numerical Data}\label{sec7.2.2}
This particular trend emerged from the necessity to detect sarcasm arising from numbers. In cases like “Love driving 3 hours to work every day” or “Started the day with 22\% charge on my phone. Today's gonna be great!”, understanding the role of numbers is crucial for detecting the underlying sarcasm. Kumar et al. \cite{kumar2017having} used several deep learning approaches in their study aside from some rule-based and machine learning based approaches to tackle such cases. Their deep learning based CNN-FF (CNN followed by Fully Connected Layer) model produced the best result.  Dubey et al. \cite{dubey2019numbers} also proposed some deep learning architectures to deal with sarcasm expressed through numbers. One is a CNN-FF model and another is made of an attention network. Both showed improvement over past works, with CNN-FF surpassing the attention network with a F1-score of 0.93 opposed to 0.91.    

\subsection{Transformers}\label{sec7.3}
Another trend that resurfaced in sarcasm detection recently is the use of transformers. A transformer is a deep learning model which uses a self-attention mechanism to weigh the input data according to a learned measure of relevance\footnote{\url{en.wikipedia.org/wiki/Transformer_(machine_learning_model)}}. Although transformers were introduced in 2017 by Vaswani et al. \cite{vaswani2017attention}, the utilization of transformers for detecting sarcasm is fairly recent. Potamias et al. \cite{potamias2020transformer} was the first to employ a methodology based on unsupervised pre-trained transformers. Their proposed methodology RCNN RoBERTa applies Recurrent CNN on a pre-trained transformer-based network architecture and outperformed state-of-the-art approaches like BERT, XLnet, ELMo, and USE. After this, the trend of employing pre-trained language models for sarcasm classification problems only grew. Transformers have been used in numerous applications in this domain. We will be taking a deeper look at two of such cases of transformer-based architectures: 1. utilizing conversational context and 2. using Multiple Modals.

\subsubsection{Utilizing Conversational Context}\label{sec7.3.1}
Transformer-based models utilize context for their sarcasm detection mechanism \cite{gregory2020transformer,javdan2020applying,avvaru2020detecting,dong2020transformer}. Among the other types of context, conversational context has been used more frequently in transformer-based architectures in the studies we have come across. Such scenarios can be seen prominently throughout the 2nd Workshop at the Figurative Language Processing 2020 shared task (FigLang2020\footnote{\url{https://competitions.codalab.org/competitions/22247}}) where conversational context was included in the provided Twitter and Reddit datasets. The contextual information included both the immediate context (i.e., the previous dialogue turn only) and the full dialogue thread, should it be available. Since transformers are effective in tracking relationships among sequential data, many teams opted for an architecture that makes use of transformer layers in their model to utilize the provided context. In the FigLang2020 shared task, Dong et al. \cite{dong2020transformer} presented a model that used deep transformer layers which utilized the entire conversational context. This resulted in 3.1\% and 7.0\% improvements over the provided Ghosh et al. \cite{ghosh2018sarcasm} baseline of 0.67 and 0.6 F1 scores for the Twitter and Reddit datasets respectively. In the same shared task, Lee et al. \cite{lee2020augmenting} presented an architecture where transformer encoder is stacked with BiLSTM \cite{schuster1997bidirectional} and NeXtVLAD \cite{lin2018nextvlad}. They generated new training samples using a data augmentation technique, CRA (Contextual Response Augmentation), which utilizes the conversational context of the unlabeled dataset and also explored multiple context lengths with a context ensemble method. With this, they observed a significant increase in F1 scores of 0.931 and 0.834 in the Twitter and Reddit datasets respectively. 

\subsubsection{Using Multiple Modals}\label{sec7.3.2}
Transformer-based models are recently being used along with image encoders in multimodal sarcasm detection. The Hierarchical Fusion Model of Cai et al. \cite{cai2019multi} is one of the first notable work that incorporated multiple modals in their sarcasm detection architecture. As mentioned before, their architecture first takes the text, image and image-attribute features, then reconstructs and fuses them. However, reconstructing from features meant leaving out certain details. As transformer-based models were introduced, pretraining models on image-text data became popular \cite{lu2019vilbert,lu202012,alberti2019fusion}. But Wang et al. \cite{wang2020building} argued that instead of only pretraining them on image-text data, BERT should be pretrained on much larger text data and so should ResNet on image data. In light of this, they developed an architecture that employs pretrained BERT and pretrained ResNet directly, without further pretraining, and then built a bridge between the two. This offers flexibility in the architecture as any transformer-based model can be swapped with BERT and ResNet can be changed to any other visual model as well. Pan et al. \cite{pan2020modeling} also tried to improve upon the Cai et al. \cite{cai2019multi} architecture. Pan et al. \cite{pan2020modeling} focused on the incongruity between image, text, and hashtags. They created a relation between text and hashtags with the help of a co-attention matrix and performed text image matching by using BERT for text and ResNet-152 for image encoding respectively. Then the results of these two relations were compared. This integration of transformer-based encoders with image encoders significantly improved the performance of multimodal sarcasm detection.

\section{Issues and Challenges}\label{sec8}
In this section, we discussed the common issues and challenges in the domain of sarcasm detection in social media platforms. We also included how and to what extent these challenges have been dealt with in previous studies.

\subsection{Language of Social Platforms}\label{sec8.1}
The words used in social media are not always limited to the dictionaries and neither is the language only confined to grammar. These non-vocabulary words and non-grammatical context make detecting sarcasm in platforms like Twitter and Reddit difficult. Moreover, hyperbolics and emblematics are often found in sarcastic sentences. Multilingual texts have also become common in social media. As a result, models trained on a single language may not always be sufficient.

\subsection{Handling Dataset Skews}\label{sec8.2}
One of the issues that arise while training models to classify sarcasm, is the scarce presence of sarcasm in real-world social media. Sarcastic sentences don't appear as frequently in everyday scenarios of social platforms as they do in a balanced dataset (Ptáček et al. \cite{ptavcek2014sarcasm}). This is why many studies take into account such imbalance when designing the framework for detecting sarcasm. Abercrombie et al. \cite{abercrombie2016putting} addressed the issue of the rarity of sarcastic tweets in Twitter and dealt with this by creating a novel imbalanced dataset of 2,240 manually annotated Twitter conversations where 20\% of the texts are sarcastic. Poria et al. \cite{poria2016deeper} also proposed an architecture that tries to work around such skews. They trained their CNN models with both balanced and imbalanced datasets. These balanced and imbalanced datasets\footnote{\url{https://liks.fav.zcu.cz/sarcasm/}} were originally introduced by Ptáček et al. \cite{ptavcek2014sarcasm}. Ptáček et al. \cite{ptavcek2014sarcasm} used Twitter Search API and Java Language Detector\footnote{\url{https://code.google.com/archive/p/jlangdetect/}} for collecting the datasets. They also collected tweets with the \#sarcasm hashtag. The balanced dataset contained 50,000 sarcastic and 50,000 non-sarcastic tweets. As sarcastic tweets are less frequently used, they trained using an imbalanced dataset of 25,000 sarcastic and 75,000 non-sarcastic tweets as well.

\subsection{Variable Context Length}\label{sec8.3}
In recent years, it has become very common to use contextual information to decipher the underlying sarcasm on social media sites. In fact, transformer-based studies heavily rely on contextual information. However, researchers are yet to come to a conclusion on how much context is actually necessary. Some studies tried to find the optimal length of conversational context while others opted for a different route by utilizing the whole conversation with variable lengths. For example, Lee et al. \cite{lee2020augmenting} made use of the entire conversational context for their transformer-based model. To deal with multiple context lengths, they needed to deploy a context ensemble method. Studies that tried to find the optimum length, came up with different conclusions. As can be seen in the Shared Task section (Section \ref{sec6.5}), Baruah et al. \cite{baruah2020context} found that including the latest uttered sentences improved the performance slightly. Experiments of Avvaru et al. \cite{avvaru2020detecting} showed that the previous 5 to 7 recent sentences in the Reddit and Twitter dataset gave the best result while Jaiswal et al. \cite{jaiswal2020neural} thinks the latest 3 utterances gives the most accurate classification of sarcasm. Thus, despite the fact that context is a crucial component of sarcasm detection, the variable length continues to make the task difficult for the researchers.

\subsection{Need of General Knowledge Beyond the Context Provided}\label{sec8.4}
Although many studies have used conversational, authorial, topical, and visual cues as contexts, there are still instances where such contexts fail to detect sarcasm. In such cases, general common sense knowledge is necessary. According to Kreuz et al. \cite{kreuz2002asymmetries}, the likelihood of sarcasm is linked to the amount of information that the speaker and audience have in common, including their own knowledge as well as knowledge of the outside world. Misra et al. \cite{misra2019sarcasm} argued that if the model has no understanding of common sense knowledge then it would just pick up some discriminative lexical cues for sarcasm upon training instead of actually understanding what makes a sentence sarcastic. Bosco et al. \cite{bosco2013developing} discussed knowledge-based approaches and suggested using paradigms for semantic annotation that rely on resources like SenticNet\footnote{\url{http://sentic.net}}, which infers both conceptual and emotional information related to opinions expressed in natural language. Tsai et al. \cite{tsai2013building} proposed a commonsense knowledge based approach of an integrated value propagation method for developing a concept level sentiment dictionary. Although in recent years, machine learning and deep learning based approaches have become popular, we did not encounter many studies that took general or common sense knowledge into consideration. Misra et al. \cite{misra2019sarcasm} suggested creating a hybrid neural network architecture that is trained on a News Headlines dataset\footnote{\url{https://www.kaggle.com/rmisra/news-headlines-dataset-for-sarcasm-detection}}, created from TheOnion\footnote{\url{https://www.theonion.com/}} and HuffPost\footnote{\url{https://www.huffpost.com/}}, which provides detailed contextual information on the sarcastic news. This approach can be further utilized by using it as a pre-computation step for learning general cues for sarcasm. After which the learned parameters can be transferred and further tuned on social media based datasets. We found Jamil et al. \cite{jamil2021detecting} to have used such multi-domain approach in their study. They proposed a CNN-LSTM model which is trained on the News Headlines and Twitter dataset and validated on News Headlines, Reddit and Sarcasm V2 dataset\footnote{\url{https://nlds.soe.ucsc.edu/sarcasm2}} (Oraby et al. \cite{oraby2017creating}).       

\subsection{Lack of Real-time Sarcasm Detection}\label{sec8.5}
Analyzing real-time data is a challenging task. Yet, many NLP-based research using real-time data have been carried out. Sentiment analysis using data acquired in real time has also been the subject of several studies \cite{wang2012system,calais2011bias,goel2016real,das2018real}. But there haven't been many real-time data analysis research on the subject of sarcasm detection. One particular study on this domain using real-time data was performed by Bharti et al. \cite{bharti2016sarcastic}. Using real-time Twitter data, Bharti et al. \cite{bharti2016sarcastic} devised a big data strategy for sarcastic sentiment detection. Utilizing the Hadoop framework, they used Apache Flume\footnote{\url{http://www.flume.apache.org/}} and Hive (Thusoo et al. \cite{thusoo2018hive}), respectively, to capture and process the data. The data was stored into Hadoop Distributed File System (HDFS)\footnote{\url{https://hadoop.apache.org/docs/r1.2.1/hdfs_design.html}, \url{https://www.ibm.com/topics/hdfs}} and for the actual detection of sarcastic tweets, the MapReduce (Condie et al. \cite{condie2010mapreduce}) programming model was used. They proposed three approaches: the Parsing Based Lexicon Generation Algorithm (PBLGA), the Interjection Word Start (IWS), and the Positive Sentiment With Antonym Pair (PSWAP). It took 11,609 seconds and 4,147 seconds, respectively, for all three combined approaches to analyze sarcasm in 1.4 million tweets without and with the Hadoop framework. This equates to a 66\% reduction in processing time under the Hadoop framework. However, following this one, we did not come across any more studies on real-time sarcasm detection. But given the volume of data generated these days in social networking sites, sarcasm detection in real time is becoming less of a choice and more of a necessity. And considering the correlation between sarcastic sentence detection and sentiment analysis, the significance of this task increases even more.

\section{Conclusion}\label{sec9}
In recent years, research on sarcasm detection, particularly on social media data has been on the rise, prompting us to conduct a survey on these works. In this paper, we reviewed various methodologies and approaches towards sarcasm detection, the trends observed over the years, datasets and features used, as well as the issues and challenges that still prevail in this field. Comparative analyses on datasets, features and approaches in tabular forms have also been provided. We observed a recent trend of using deep learning and transformer based approaches. Additionally, we noticed an increase in the use of contextual features such as conversational context, author information etc. Use of visual data and multimodal sarcasm detection are also fairly recent trends as observed in our survey. Lastly, we look into the issues and challenges encountered in these studies, for instance the difficulty in grasping the language used on social platforms, handling contextual features, lack of sarcasm detection in real-time and such. To provide a more detailed and thorough understanding of these studies, a chronological representation of the trends in sarcasm detection and comparative analysis tables are given.


\bibliography{sn-article}


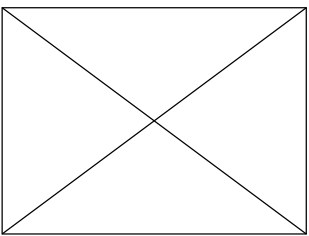
\end{document}